\documentclass[sigconf]{acmart}

\renewcommand\footnotetextcopyrightpermission[1]{} 
\usepackage{blindtext, graphicx}
\include{placeins}
\usepackage{placeins}
\usepackage[utf8]{inputenc}
\usepackage{tabularx}
\usepackage{multirow}
\usepackage{listings}
\usepackage{todonotes}
\usepackage{textcomp}

\lstdefinestyle{mystyle}{
    basicstyle=\footnotesize,
    breakatwhitespace=false,         
    breaklines=true,                 
    captionpos=b,                    
    keepspaces=true,                 
    numbers=left,                    
    numbersep=5pt,                  
    showspaces=false,                
    showstringspaces=false,
    showtabs=false,                  
    tabsize=2
}
\begin{document}

\title[Autonomous Configuration of Network Parameters]{Autonomous Configuration of Network Parameters in Operating Systems using Evolutionary Algorithms}


\author{Bartosz Gembala}

\affiliation{%
  \institution{Department of Computer Science\\
  Oslo Metropolitan University}
  \city{Oslo}
  \state{Norway}
}

\author{Anis Yazidi}

\affiliation{%
  \institution{Department of Computer Science\\
  Oslo Metropolitan University}
  \city{Oslo}
  \state{Norway}
}
\email{anisy@oslomet.no}

\author{H\aa rek Haugerud}

\affiliation{%
  \institution{Department of Computer Science\\
  Oslo Metropolitan University}
  \city{Oslo}
  \state{Norway}
}

\author{Stefano Nichele}

\affiliation{%
  \institution{Department of Computer Science\\
  Oslo Metropolitan University}
  \city{Oslo}
  \state{Norway}
}

\begin{abstract}
By default, the Linux network stack is not configured for high-speed large file transfer. The reason behind this is  to save memory resources. It is possible to tune the Linux network stack by increasing the network buffers size for high-speed networks that connect server systems in order to handle more network packets. However, there are also several other TCP/IP parameters that can be tuned in an Operating System (OS). In this paper, we leverage Genetic Algorithms (GAs) to devise a system which learns from the history of the network traffic and uses this knowledge  to optimize the current performance by adjusting the parameters. This can be done for a standard Linux kernel using sysctl or /proc. For a Virtual Machine (VM), virtually any type of OS can be installed and an image can swiftly be compiled and deployed. By being a sandboxed environment, risky configurations can be tested without the danger of harming the system.
Different scenarios for network parameter configurations are thoroughly tested, and an increase of up to 65\% throughput speed is achieved compared to the default Linux configuration.

\end{abstract}


\maketitle

\section{Introduction}
There is a large number of parameters which can be set and changed when configuring the network of a server and this leads to a huge number of possible configurations. Genetic algorithms (GAs) \cite{smith,genintro,zengen} are most often used to solve search and optimization problems \cite{goldberg}. In this paper, we will resort to those algorithms to discover the most efficient set of parameter combinations for the given problem. The premises of GAs when applied to network configuration is that better configuration sets will have a higher chance of survival and being optimized through multiple rounds (generations) of selection, crossover and mutation. GAs are being used for the sake of efficiently searching through the immense number of available combinations, when brute force testing of all the different combinations is unfeasible.

A major part in choosing the right network configurations is to find the most relevant network parameters. The goal of the optimization may be to prioritize the throughput for fast transfer of data but giving latency some level of importance might also be relevant. This choice depends on the nature of the service at hand. The default network settings are usually generic and static regardless of the traffic going through and hence they do not generally optimize the network performance for all kinds of network load. A correct configuration setting can fully utilize the system resources and hence lead the system to the best Quality of Service (QoS) for properties such as short request response time and high throughput.

In this paper, we will be performing various tests both on physical machines and Virtual Machines (VMs) to compare the diverse options which can be derived for different purposes. Physical machines can give us better performance, but there are many benefits of using VMs. VMs enable multiple workloads to be consolidated on less servers and safe isolation of co-located workloads. This improves resource utilization, reduces idle power costs and makes it possible to test risky configurations.

There are many ways to increase throughput, for example the quality can be sacrificed \cite{goldberg}, or various parameters can be tuned \cite{netcards,tcptune}. There are some  systems that improve simply by changing the different configurations \cite{boostperf,autotune} and tuning them in accordance to what one wants to achieve. The main problem is to tune those different parameters dynamically in accordance to the payload which is received.

At this juncture, we shall review some related research.
GAs have been applied in the literature to  tighten security \cite{adam} via manipulating  the configuration parameters of Apache 2.0.

The work reported in \cite{apachecon} presents learning through Reinforcement Learning (RL) to improve different configuration parameters. It uses the RL approach for autonomic configuration and reconfiguration of multi-tier web systems. It is able to adapt performance parameter settings not only to the change of workload, but also to the change of VM configurations. The approach is evaluated using the TPC-W benchmark on a three-tier web-site hosted in a Xen-based VM environment \cite{apachecon}. The results of the experiments demonstrate that the approach can auto-configure the web system dynamically in response to the change in both workload and VM resources. The focus is on Apache parameters, but the auto-configuration part is relevant to what we are trying to achieve herein.

Each generation of network cards has different features, and if not fully configured the network performance might become a severe bottleneck. However for Linux, where the operating system runs on various types of machines, the default configurations are not tuned to for 10 Gbit/s network cards, or 1 Gbit/s in our case. \cite{netcards} describes the basic settings that can be changed in a Linux environment in order to maximize the throughput speed.

The performance of the receive side TCP processing has usually been overruled by "per-byte" operations, such as check-summing and copying. But as the architecture in modern processors has changed, "per-packet" operations are becoming the main source of overhead \cite{tcpreceive}. Two optimization techniques are represented to improve the receive side TCP performance. A similar benchmark for testing the TCP streaming receive throughput with netperf is used. Results of another study \cite{tcpoverhead} shows that TCP is not the source of overhead often observed in packet processing, and it could support a lot higher speeds if correctly implemented.

The reminder of this article is organized as follows.
In Section \ref{sec:Genetic algorithm}, a short overview of the principles of GAs is given. Section \ref{Sec:Approach} provides an overview of our solution.
Section \ref{sec:Experiments} gives an overview of the experimental set-up and reports the experimental results.
Section \ref{Sec:Conclusion} concludes the article and gives an overview of future research directions worth exploring.


\section{Genetic algorithms (GAs)}
\label{sec:Genetic algorithm}

GAs are inspired by the process of natural selection and are a version of Evolutionary Algorithms (EAs) \cite{genintro}. They are most often used to solve search and optimization problems by generating high quality solutions using biologically inspired operators such as selection, crossover and mutation. The common underlying idea behind this is: given a population of individuals within some environment that has limited resources, competition for those resources causes natural selection (survival of the fittest). This results in a rise of fitness of the population. Then based upon fitness of the candidates the better candidates are chosen to seed the next generation. After a given  amount of generations the GA will come to a point where it is not getting considerably fitter offspring, that is when the GA stops. GAs have also been successfully applied to many theoretical optimization problems \cite{goldberg} and several industrial applications \cite{davis1991handbook}.

\subsection{Representation}

In genetics a genotype is the part of the genetic makeup of a cell, and phenotype is what determines its characteristics.

	\begin{figure}[th!]
		\includegraphics[width=0.48\textwidth]{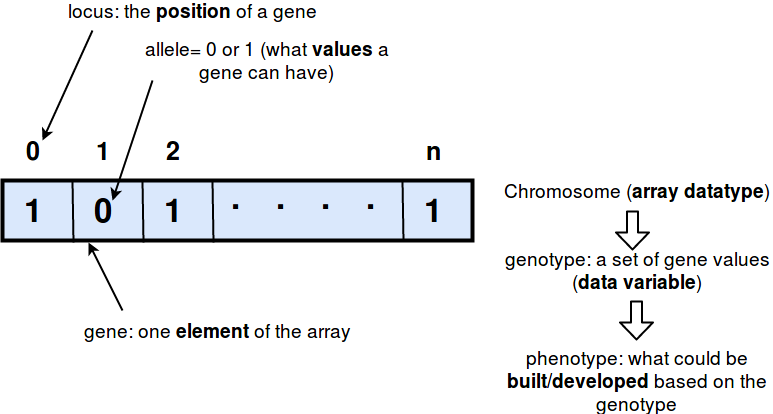}
		\caption{Genotype Representation}
		\label{fig:genrep}
	\end{figure}
	
By using GAs we can represent our configurations on a genetic level, as a genotype as seen in Figure \ref{fig:genrep}. 
The array data type is called a chromosome, and each chromosome consists of multiple genes, while the possible values for each gene are called alleles. A programmer can represent all individuals in a population in many different ways, some of them being; binary, integer, permutation, real-valued or floating-point. In our algorithm each chromosome is a representation of a parameter combination, whereas a gene is a single parameter among many.

\subsection{Life cycle}

The genetic operator is an operator used in GAs that leads the algorithm toward a solution. The three types of operators are selection, crossover and mutation operators.
These operators contribute to crossing existing solutions into new solutions (crossover), diversifying the population (mutation) and selecting the best solutions for every generation (selection).

 \begin{figure}[th!]
	\includegraphics[width=0.48\textwidth]{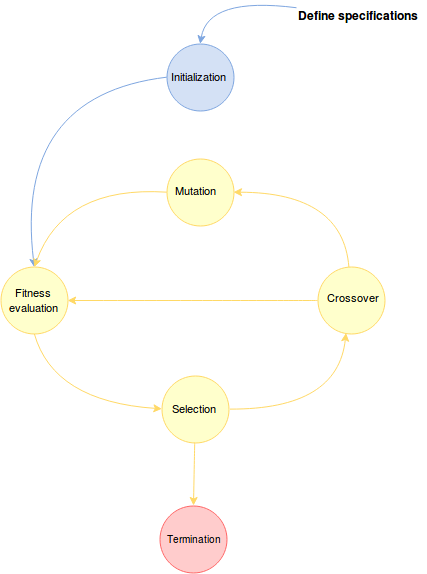}
	\caption{The algorithmic life cycle, representing the different processes taking place from initialization of a population until its termination.}
	\label{fig:flow1}
\end{figure}


In Figure \ref{fig:flow1} a population goes through evaluation and selection, this operation gives better individuals stronger preference and allows them to pass their genes on to the next generation. How good an individual is depends on their fitness, and there are different methods to choose from, such as fitness proportional selection, ranking selection, tournament selection etc. For every method there is a different criteria of what being fit is. The fittest parents are chosen and a crossover operation is performed on parents based on the crossover method. Crossover or recombination is an operation that merges information from two parent genotypes into one or two offspring genotypes. Crossover is a stochastic operator where the algorithm decides what part of each parent will be combined. The main idea behind this is to partner up two different individuals with desirable features to create an offspring which combines those features. This has been done over millennia to plants and by livestock breeders to produce species that give higher yield or have other desirable features \cite{smith}.

The next operator in the cycle is mutation, which contributes to diversity from one generation to another in the GA. The goal is to change one or multiple gene values to something different. The mutation probability is defined in the algorithm. At the end of the cycle new parents will be chosen depending on their fitness, resulting in new fitter individuals merging with the population while the less fit individuals are removed.


\section{Approach}
\label{Sec:Approach}

	\subsection{Selection of configuration parameters}

	Netperf is used for measuring the network speed. It is a software application that can be used to measure different aspects of network performance. It supports Unix domain sockets, but is mostly used on bulk data transfer request/response using TCP or UDP and Berkeley Sockets interfaces \cite{netperf}. It provides numerous predefined tests.

	Just like in the case of cluster-based web service performance, the performance improvement can not easily be achieved by tuning individual components \cite{clusterserv}, there is no single universal configuration that is good for all workloads. Therefore the genetic algorithm is used to find the optimal configuration setup for a given payload which is going through the server. 
   
    There are many relevant parameters described in \cite{tcptune,netcards} and some of them are:

	\begin{itemize}
		\item \textbf{Jumbo Frames -}  Jumbo frames are Ethernet frames with a payload of more than 1500 bytes, which is the standard limit \cite{jumbo}. Per definition, jumbo frames can carry up to 9000 bytes. In combination with Gigabit Ethernet switches, network interface controllers (NICs) can support frames bigger than the default, making jumbo frames possible. The network parameter determining the frame size is the Maximum Transfer Unit (MTU).
		
		\item \textbf{Multi streams -} The throughput of a network depends on the type of traffic that is flowing through the wire. The number of streams is important, that is, how many end to end socket connections which are established and uses the network at the same time\cite{tcptune}. The Netperf tool has a mode where one can use up to 8 active streams. 
		
		\item \textbf{Transmission queue size -} The transmission queue is a buffer which holds packets that are scheduled to be sent to the card. If one wants to avoid the packet descriptors being lost, the size of the buffer should be tuned. The default size of 1000 packets can be too small.
		
		\item \textbf{TX Checksum -} This is a checksum offload parameter that when set asks the card to compute the segment checksum before it is sent. When this is enabled, the kernel enters a random value in the checksum field of the TCP header and leaves it to the network adapter to calculate the correct checksum. 
		
		\item \textbf{TCP Segmentation Offload - } This is a parameter which may be used to reduce the CPU overhead when running TCP/IP. The task of splitting a large chunk of data into TCP segments, which normally is done by the OS, is handed over to the NIC, which splits the data into segments and add fills in the TCP header fields correctly.
		
		\item \textbf{Large Receive Offload - } If this parameter is set and supported by the NIC, it  combines multiple Ethernet frames received into one large frame, offloading this work for the OS and the CPU. It is a technique which increases inbound throughput of high-bandwidth network connections. There are two types of LRO, one that is usually turned on by default and another which is specific for the given device driver. The last one yields much better results as it accumulates the frames inside the NIC.
		
		\item \textbf{Generic Segmentation Offload (GSO) - } It has been observed that a lot of savings in TSO come from traversing the network stack once rather than multiple times\cite{gso}. GSO is, like TSO, only effective if the MTU is around the default value of 1500.
		
		\item \textbf{TCP Window Scaling - } Is an option that can increase the size of the receive window allowed in TCP beyond its default value of 65535 bytes. The throughput is limited by two windows, the receive and the congestion window. The receive window tries not to go past the limit of the receiver to process data, while the congestion window tries not to breach the limit of the network (congestion control).
		
		\item \textbf{TCP Timestamp - } Enabling this can provide a more accurate round trip time measurement, but it also adds an overhead to the throughput and CPU usage. This option should be disabled if one wants to increase the speed.
		
		\item \textbf{Memory - } There are three parameters, exemplified by \textit{net.ipv4.tcp\_mem = 287121 382828 57424}, which define how the kernel should manage the memory usage of the TCP stack. The first value tells the kernel that below this number of memory pages, no restriction on memory should be imposed. The second value defines the starting point at which the kernel should start pressuring memory usage down. The final defines the maximum amount of memory pages for all sockets.
		
		\item \textbf{Read and Write Memory - } two parameters control the read and write memory buffers, \textit{net.ipv4.tcp\_rmem} and \\ \textit{net.ipv4.tcp\_wmem}. Read memory takes care of the size of the receive buffer used by TCP sockets, while write memory adjusts the amount of memory reserved for send buffers. Each have three values, exemplified by  \textit{net.ipv4.tcp\_rmem = 4096 87380 16777216}. The first value tells the kernel the size of the minimum receive buffer for a TCP connection which is allocated to a TCP socket. The second is the default receive buffer and the third is the maximum receive buffer. Similarly   \textit{net.ipv4.tcp\_wmem} determines the size of the TCP send buffer of TCP sockets in use.
	\end{itemize}

\subsection{The fitness function}
There are many ways to measure network performance as every network is different in nature and design. And how important the various metrics are depends on what the network is used for, for instance what kind of service the servers involved provide. A few of the most used metrics are the following:

\begin{itemize}
	    \item \textbf{Throughput} is usually measured in bits per second and is the amount of data which is transmitted through a network per unit time. 
	    
	    \item \textbf{Latency} is the response time, the delay between sender and receiver, expressing how fast the signal travels from one end to another.
	    
	    \item \textbf{Bandwidth} determines that maximum possible throughput that is achievable, one can not send more data than the specified bandwidth size.
	    
	    \item \textbf{Jitter} is the variation in packet delays, normally an undesired deviation.
	    
	    \item \textbf{Packet loss} is when packets of data transmitted through a computer network fail to reach their destination.
	    
	    \item \textbf{Quality of Service} (QoS) is a measure of the overall performance of a service as seen by the users. 
\end{itemize}

We decided to only focus on throughput, i.e. the fitness function equals the measured throughput of the network for a given set of parameters. However, a few tests where made checking whether the change of parameters leading to higher throughput also affected the latency. 
The tests measuring the fitness function are applied to TCP streams, having in mind traffic such as large video files being transferred in a network. 

The tools used in this paper to measure throughput are iPerf \cite{iperf} and netperf\cite{netperf}. iPerf is a tool to measure the maximum achievable bandwidth on a network. There are various parameters which can be customized, buffers and protocols such as TCP and UDP. iPerf measures bandwidth, packet loss, transfer size and other parameters. It has a client-server functionality and can generate data streams in order to measure the throughput, in one or two directions. Netperf is another software application that measures network throughput between two hosts. It has a variety of tests to measure unidirectional data transfer and request/response performance. In order to measure latency, ping and hping3 are used, simply measuring the round-trip time for messages sent from the originating host to the destination computer. An Internet Control Message Protocol (ICMP) echo request packet is sent waiting for an ICMP echo reply.

\subsection{The algorithm}



Figure \ref{fig:flow1} shows a simplified flowchart for the entire algorithm. It starts with the initialization of the population. Each parent is a chromosome that consists of a set of parameters, whereas a gene is one of these parameters. Several sources\cite{netcards, tcptune} were studied in order to select adequate parameters. The list of 14 different parameters shown below were used for initial experiments and this list was at a later stage extended to a total of 27 parameters. For every run of the algorithm, each parameter from the list can be extracted, but it is the algorithm that decides whether all the parameters will be used or just a selected few, based on the fitness. 
\lstset{basicstyle=\small}
\begin{lstlisting}[ caption=Parameter list,numbers=none]
sysctl -w net.ipv4.tcp_mem=;'287121 382828 574242';'16777216 16777216 16777216'
sysctl -w net.ipv4.tcp_rmem=;'4096 87380 6291456';'8192 873800 16777216'
sysctl -w net.ipv4.tcp_wmem=;'4096 16384 4194304';'8192 873800 16777216'
sysctl -w net.ipv4.tcp_moderate_rcvbuf=;0;1
sysctl -w net.ipv4.tcp_no_metrics_save=;0;1
sysctl -w net.ipv4.tcp_timestamps=;0;1
sysctl -w net.ipv4.tcp_window_scaling=;0;1
sysctl -w net.ipv4.tcp_sack=;0;1
sysctl -w net.core.wmem_max=;212992;16777216
sysctl -w net.core.rmem_max=;212992;16777216
sysctl -w net.core.rmem_default=;212992;412992
sysctl -w net.core.wmem_default=;212992;412992
sysctl -w net.core.netdev_max_backlog=;1000;5000
ifconfig eno2 mtu ;1500;2700
\end{lstlisting}

At the end of each line, the minimum and maximum values used for each parameter is shown. When initializing the first generation, the genetic algorithm chooses a random value between the minimum and the maximum value.

When all the parents have been initialized and placed in the population pool, the next step is to measure the fitness for every member of the population. One by one, each parent is tested by applying all its network parameter genes to 
the server and then performing a netperf test, normally ten seconds long. 
The throughput speed recorded by netperf is then stored for each parent as its fitness function.


The next phase of the genetic life cycle is the selection. The percentage of the population that should be disposed of for every generation is one of the configuration parameters of the algorithm. In the experiments presented in this paper, a value of 10\% was used. This means that 10 percent of the worst fit individuals is removed in every generation. 

After the selection, the population is restored by selecting the best 10\% of the parents for generation of the same number of offspring. All the pairs of this selection of parents makes a pair of children by going through a crossover process where there is a given chance for a switch of genes between the two parents leading to their offspring. In our experiments this  probability was set to 50\%. 

In the final phase a mutation may occur to any of the specimen, the probability was set to 16\% in our case. In this process, if a parent is selected for mutation, one gene of the chromosome is given a new random value.    

This cycle repeats a number of times. In a real life system this would go on forever, but in our algorithm we terminate the loop after an number of iterations, mostly determined by how stable the results are.

%

%
%

%

\section{Experiments}
\label{sec:Experiments}
\subsection{Experimental setup}

\begin{figure}[th!]
        \center
    	\includegraphics[width=0.48\textwidth]{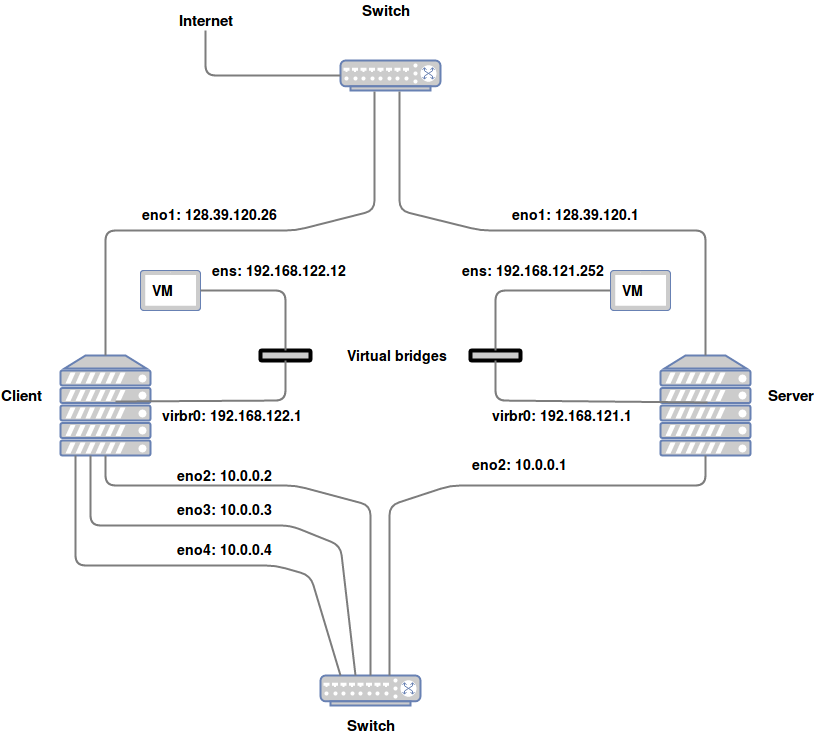}
    	\caption{The topology for the setup using two physical machines with 1 Gigabit ethernet ports. The machines are connected through a switch which forwards the data traffic of the experiments.}
    	\label{fig:topology3}
    \end{figure}

The experimental setup is shown in Figure \ref{fig:topology3}. There are two physical servers running Ubuntu 16.04 which are connected through a switch to the internet. Another switch connects the 1 gigabit ethernet ports of the servers to each other. The specifications for these machines are listed in Table \ref{table:machspec}.

\begin{table}[!pht]
    \scalebox{0.7}{
    \begin{tabular}{ |c|c| } 
    \hline
    OS & Ubuntu 16.04 xenial\\ 
    \hline
    CPU & 2x Intel Xeon CPU E5530 @ 2.394GHz\\ 
    \hline
    NICs & Broadcom Corporation NetXtreme II BCM5709 Gigabit Ethernet x4 \\
    & Intel Corporation 82576 Gigabit Network Connection x4 \\
    \hline
    RAM & 290MiB / 24098MiB \\
    \hline
    Disk space & 151 GiB \\
    \hline
    \end{tabular}
    }
    \caption{The machine specifications of our servers}\label{table:machspec}
    \end{table}

Both of these machines have four Network Interface Controllers (NICs). The payload of the experiments is generated using iperf. On the computer running the clients, three of the four NICs available are bound to an iperf process, and the last one is used for ssh access. In this way they can separately generate traffic. Two VMs are included in the topology for the purpose of testing the algorithm and its behaviour on virtual machines.

\subsection{Test scenarios}
Just as for cluster-based web service performance, the performance improvement can not easily be achieved by tuning individual components\cite{clusterserv}. There is no single universal configuration that is best for all workloads. Therefore, the genetic algorithm is used to find the optimal configuration setup for a given payload on the server. The different parameters used are shown in Table \ref{table:newpams} and there are 27 of them. If all the parameters were binary, there would be $2^{27}$ possible combinations. More than a 100 million combinations would evidently be impossible to test one by one in order to find the best network configuration. Additionally some of the parameters have thousands of possible values, making the total number of combinations immense and the need for a smarter algorithm obvious.

\begin{table}[!pht]

    \center
    
	
    \begin{tabular}{ |c|c|} 
    \hline
    Type & Parameters \\
    \hline
    \multirow{3}{4em}{TCP - net.ipv4}
    & tcp\_mem \\ 
    & tcp\_rmem  \\ 
    & tcp\_wmem \\ 
    & tcp\_moderate\_rcvbuf \\
    & tcp\_no\_metrics\_save \\
    & tcp\_timestamps\\
    & tcp\_window\_scaling \\
    & tcp\_sack\\
    & tcp\_tw\_reuse \\
    & tcp\_keepalive\_probes \\
    & tcp\_keepalive\_intvl  \\
    & tcp\_fin\_timeout \\
    \hline
    \hline
    \multirow{3}{4em}{net.core}
    & wmem\_max  \\ 
    & rmem\_max\\ 
    & rmem\_default  \\ 
    & wmem\_default  \\
    & netdev\_max\_backlog \\
    & bpf\_jit\_enable  \\
    & dev\_weight  \\
    & rps\_sock\_flow\_entries  \\
    & optmem\_max  \\
    & somaxconn  \\
    & busy\_read \\
    & busy\_poll  \\
    & tstamp\_allow\_data  \\
    \hline
    \hline
    MTU & mtu\\ 
    Txqueuelen & txqueuelen \\
    \hline
    \end{tabular}
    \caption{Listing of all the different parameters used in the experiments}
    \label{table:newpams}
    
\end{table}

Two different tests will be performed in this topology. In the first one the throughput speed between our physical machines while we load them with traffic is tested. In the second test the same is done using virtual machines. The performance is tested both when changing the parameters directly on the VMs as well as on the host machine. It is a way to establish whether changes on just the VMs actually have as much effect as changes on the host machine they are operating on, and also to see to what extent changing parameters on the host machine affects the performance on the VMs. The two scenarios that are to be performed are as follows:

\begin{itemize}
    \item Configurations changed directly inside the VM
    
    \item Configurations changed on the host machine (server)
\end{itemize}



For all the experiments performed, the following parameters were used as input to the genetic algorithm:
\begin{itemize}
    \item Parameter number: 27 
    
    \item Generations: 40 
    
    \item Population size: 80 
    
    \item Selection probability: 10\%
    
    \item Crossover probability: 50\%  
    
    \item Mutation probability: 16\%
\end{itemize}
The same values were used for all the experiments, making it straightforward to compare the results of the various scenarios. The values where chosen after some initial tests since they yielded the best results.

\subsection{Results}
\begin{figure}[th!]
    	\includegraphics[width=0.48\textwidth]{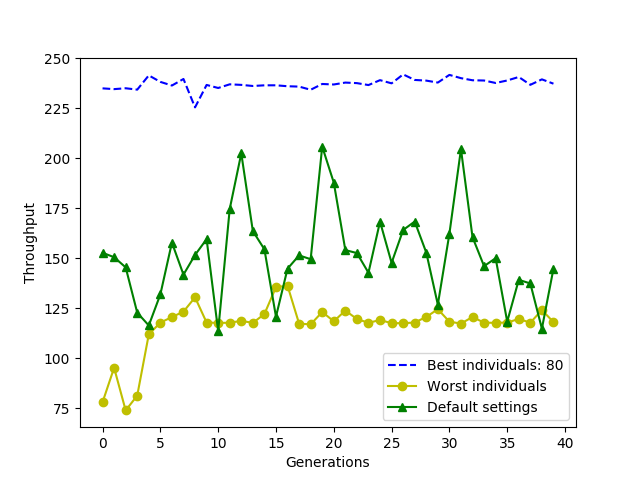}
    	\caption{The best, worst and default individuals for every generation. The best parameter settings display on average a 65\% higher throughput speed than those of the default OS settings.}
    	\label{fig:normal2}
    \end{figure}

Initial experiments showed that the algorithm was able to improve the total client-server throughput by a few percent when the fitness function was calculated based on a netperf stream which was the only one using the network and hence used the complete 1 Gbit/s bandwidth. For the main part of the experiments, and for all results shown in this section, three independent TCP connections were set up between the two physical servers. The netperf stream used for calculating the fitness function then had to compete with the other traffic. In this scenario, the genetic algorithm was able to improve the performance of the default Linux network configuration substantially.   
This can be seen in Figure \ref{fig:normal2}, where the throughput of the best individuals (the best parameter combination found for every generation) clearly are much better than the throughput obtained when using the default configurations of the Linux OS.
 
On average, the performance of the top individuals are larger by 65\% in Figure \ref{fig:normal2}, but the throughput speed does not increase substantially over the generations. This means that a good parameter combination is found already among one of the first 80 randomly initialized parents.

In order to see wether there are any drawbacks when imporoving the throughput, the latency was tested using hping3 on the  server loaded with network traffic, using the default and the optimal configurations learned from the algorithm. One of the results is shown in  Listing \ref{lst:lattest}.

\begin{lstlisting}[caption=Latency test from Client to Server, label=lst:lattest,numbers=none]
sudo hping3 -c 10000 -i u10000 10.0.0.1 -p 8000
    Fast parameter combination
        round-trip min/avg/max = 1.2/6.8/1005.8 ms
    Default
        round-trip min/avg/max = 1.3/7.0/1006.1 ms
\end{lstlisting}
Other tests also showed that the latency was not affected when increasing the throughput. 

Two types of experiments were performed on the virtual machines. One in which the network parameters of the OS of the VM itself was changed, and another where the parameters of the OS of the host machine or hypervisor of the VM was changed. In some cases changing parameters of the OS of the VM does not change the actual behaviour of the underlying hypervisor, so potentially these could lead to different results in the two cases. 

\begin{figure}[th!]
	\includegraphics[width=0.48\textwidth]{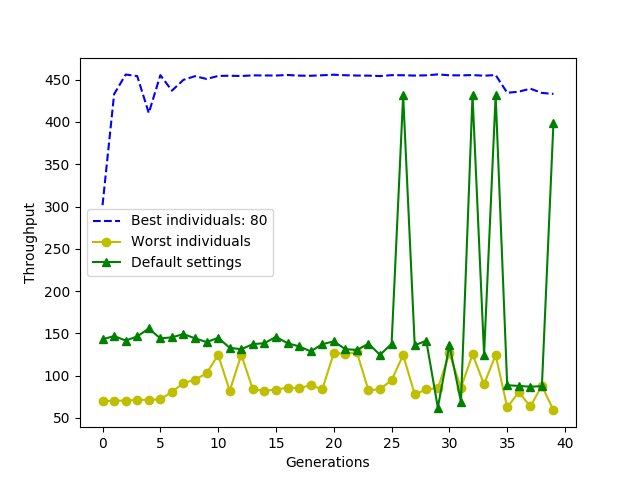}
	\caption{The best, worst and default individuals for each generation when changing the configuration of the OS of the VM.}
	\label{fig:figvmtest2}
\end{figure}

As seen in Figure \ref{fig:figvmtest2}, the throughput does increase substantially when optimizing the parameters inside the VM. Here a few generations are needed before the optimal configuration is found. It is surprising that the throughput of the traffic entering the VM-based server is so high. Some specific network configuration benefits this stream compared to the background traffic of the iperf-clients, and more research is needed to find the exact reason. However, in our case, the most important result is to see how the algorithm improves the throughput compared to the that of the default VM configuration.


In the other set of VM experiments, the default configuration of the VM itself was kept unchanged and instead the algorithm was applied to the network parameters of the OS of the hypervisor.

\begin{figure}[th!]
	\includegraphics[width=0.48\textwidth]{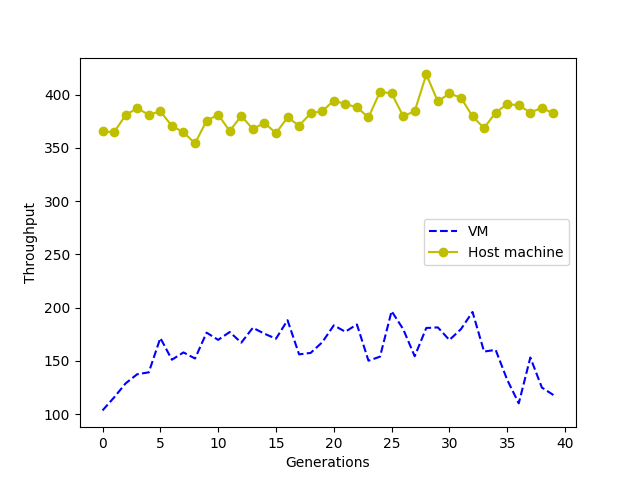}
	\caption{The average population throughput of every generation. Comparing the configuration of the VM-OS to the configuration of the host machine-OS.}
	\label{fig:vmhyp}
\end{figure}

In Figure \ref{fig:vmhyp} it can be seen that the throughput of the average population is much smaller when the configurations takes place within the VM. This might as mentioned earlier be due to the fact that the effect of changing some of the parameters of the VM-OS does not take effect in the OS of the host machine or hypervisor.

\begin{figure}[th!]
	\includegraphics[width=0.48\textwidth]{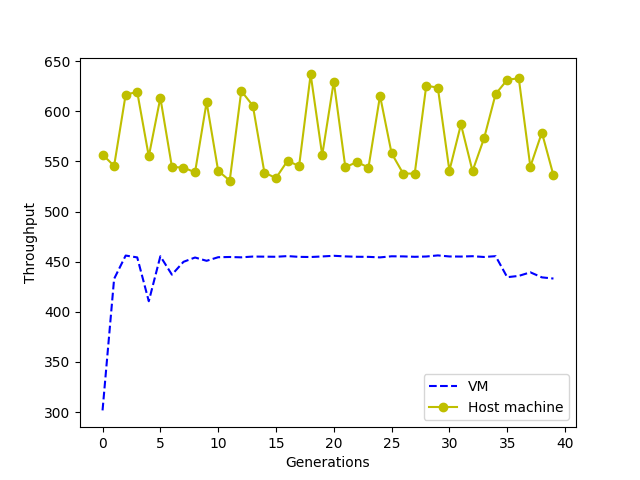}
		\caption{The best individuals of each generation. Comparing the configuration of the VM-OS to the configuration of the host machine-OS.}
	\label{fig:vmhyptop}
\end{figure}

The difference between the throughput speed in Figure \ref{fig:vmhyptop} for the top individuals is approximately 30\%, between VM and host machine. This means that by changing the parameters directly on the physical machine, we get higher performance in comparison to changing it on the VM. The speed of the top individuals in both cases are also much higher than the default settings, by 60\% on the host machine and by 180\% on the VM. 


\section{Conclusion and Future Work}
\label{Sec:Conclusion}
The main objective of this paper was to improve the throughput speed by finding the optimal set of configurations depending on the network traffic, using a genetic algorithm. After various experiments we were able to conclude that the throughput speed did improve considerably compared to that of the default settings of the Linux OS. In fact, we were able to achieve a throughput speed increase by up to 65\% for physical machines and even more for virtual machines. The fitness of the whole population raises from one generation to another, but the performances of the top individuals are quite stationary. After discovering the top individuals in the first generation, the speed does not increase significantly through more generations. This means that for the type of traffic considered here, testing 80 randomly generated configurations is enough for obtaining a very good configuration. However, for other kinds of network traffic and for traffic changing dynamically, a more advanced algorithm like the one demonstrated here, would be needed.

A not fully optimized throughput speed on a machine can be caused by many factors as the different relevant parameter values could be too low, not allowing for maximum send/receive capability. A wrong chain of configurations could also be the problem, and that is something which is impossible or very hard to detect manually due to the immense number of available combinations. Hence a genetic algorithm is being used to search for the best solution. Each set of parameters is represented as a chromosome, those chromosomes go through multiple crossover, mutation and selection processes and the most fit configuration is selected in every generation.

Going through those various operations means that our GA can find an optimized version of configurations exclusively for the type of traffic that is going through the server at a given time. An adaptive approach to change network parameters dynamically has been developed, which resulted in a significant increase of speed compared to the default settings. Our experiments furthermore indicates that the latency is not affected negatively by the optimization of the throughput. 

\subsection{Future Work}
Some experiments were done in order to test the effect of changing the percentages of several of the parameters of the genetic algorithm, 
like crossover and mutation probabilities, but this work should be extended in order to improve the algorithm. Adding more networking parameters to the configuration set could improve the results, and in addition the GA could benefit from supporting  automatic parameter extraction. In this work they were extracted from a predefined list, but every OS-version potentially has different parameters. This could be done by searching through all available parameters on the systems, saving them to a list and testing different values for each of them. Another approach that should be pursued, is to address the layer above the operating system, changing the configuration of applications running on the OS. For example, if an Apache web-server was used, the algorithm could include changing the default Apache configurations on top of the configurations of the OS, improving the overall throughput even further.


\paragraph{Ability to adjust latency and throughput ratio}
The fitness function of this work equals the throughput. A more general solution would be to include latency in the calculation of the function. Additionally, it should be possible to adjust the importance of each of these features for the server-admin which runs such an algorithm. Other metrics, like jitter and packet loss, could also be included in the fitness function.

\paragraph{A/B testing}
An A/B test could have been performed in order to find an algorithm with the best possible factors for selection, crossover and mutation. Having every instance try different probabilities, split the users and test different versions on them to see which one is preferable. One instance could have higher throughput, another have higher latency and a third one something in between.

\paragraph{Load balancing}
It could be efficient to split the traffic and send the streams to different machines so that they all could contribute in the search for an optimal solution. With the introduction of a load balancer between multiple physical machines or VMs the algorithm could be running at multiple instances at once. And if one instance finds a better parameter combination than the one already running, this combination could be distributed to all instances.



\bibliographystyle{acm}
\bibliography{NetworkBib}

\begin{thebibliography}{10}

\bibitem{apachecon}
{\sc Bu, X., Rao, J., and Xu, C.~Z.}
\newblock A reinforcement learning approach to online web systems
  auto-configuration.
\newblock In {\em 2009 29th IEEE International Conference on Distributed
  Computing Systems\/} (June 2009), pp.~2--11.

\bibitem{adam}
{\sc Buji, A.}
\newblock Genetic algorithm for tightening security.
\newblock Master's thesis, University of Oslo, 2017.

\bibitem{davis1991handbook}
{\sc Chambers, L.~D.}
\newblock {\em The practical handbook of genetic algorithms: applications}.
\newblock Chapman and Hall/CRC, 2000.

\bibitem{boostperf}
{\sc Chen, H., Jiang, G., Zhang, H., and Yoshihira, K.}
\newblock Boosting the performance of computing systems through adaptive
  configuration tuning.
\newblock In {\em Proceedings of the 2009 ACM Symposium on Applied Computing\/}
  (New York, NY, USA, 2009), SAC '09, ACM, pp.~1045--1049.

\bibitem{clusterserv}
{\sc Chung, I.~H., and Hollingsworth, J.~K.}
\newblock Automated cluster-based web service performance tuning.
\newblock In {\em Proceedings. 13th IEEE International Symposium on High
  performance Distributed Computing, 2004.\/} (June 2004), pp.~36--44.

\bibitem{tcpoverhead}
{\sc Clark, D.~D., Jacobson, V., Romkey, J., and Salwen, H.}
\newblock An analysis of tcp processing overhead.
\newblock {\em IEEE Communications Magazine 27}, 6 (June 1989), 23--29.

\bibitem{autotune}
{\sc Diao, Y., Hellerstein, J.~L., Parekh, S., and Bigus, J.~P.}
\newblock Managing web server performance with autotune agents.
\newblock {\em IBM Syst. J. 42}, 1 (Jan. 2003), 136--149.

\bibitem{iperf}
{\sc Dugan, J., Elliott, S., Mah, B.~A., Poskanzer, J., and Prabhu, K.}
\newblock iperf.

\bibitem{tcptune}
{\sc Dunigan, T., Mathis, M., and Tierney, B.}
\newblock A tcp tuning daemon.
\newblock In {\em ACM/IEEE 2002 Conference Supercomputing\/} (2002), IEEE,
  pp.~9--9.

\bibitem{smith}
{\sc Eiben, A.~E., and Smith, J.~E.}
\newblock {\em Introduction to Evolutionary Computing}, 2nd~ed.
\newblock Springer Publishing Company, Incorporated, 2015.

\bibitem{jumbo}
{\sc Feng, W.-c., Balaji, P., Baron, C., Bhuyan, L.~N., and Panda, D.~K.}
\newblock Performance characterization of a 10-gigabit ethernet toe.
\newblock In {\em Proceedings. 13th Symposium on High Performance
  Interconnects, 2005.\/} (2005), IEEE, pp.~58--63.

\bibitem{goldberg}
{\sc Goldberg, D.~E.}
\newblock {\em Genetic Algorithms in Search, Optimization and Machine
  Learning}, 1st~ed.
\newblock Addison-Wesley Longman Publishing Co., Inc., Boston, MA, USA, 1989.

\bibitem{zengen}
{\sc Goldberg, D.~E.}
\newblock Zen and the art of genetic algorithms.
\newblock In {\em Proceedings of the Third International Conference on Genetic
  Algorithms\/} (San Francisco, CA, USA, 1989), Morgan Kaufmann Publishers
  Inc., pp.~80--85.

\bibitem{netperf}
{\sc Jones, R., et~al.}
\newblock Netperf: a network performance benchmark.
\newblock {\em Information Networks Division, Hewlett-Packard Company\/}
  (1996).

\bibitem{netcards}
{\sc Leitao, B.~H.}
\newblock Tuning 10gb network cards on linux.
\newblock In {\em Proceedings of the 2009 Linux Symposium\/} (2009), USENIX
  Association.

\bibitem{tcpreceive}
{\sc Menon, A., and Zwaenepoel, W.}
\newblock Optimizing tcp receive performance.
\newblock In {\em USENIX 2008 Annual Technical Conference\/} (Berkeley, CA,
  USA, 2008), ATC'08, USENIX Association, pp.~85--98.

\bibitem{genintro}
{\sc Mitchell, M.}
\newblock {\em An introduction to genetic algorithms}.
\newblock MIT press, 1998.

\bibitem{gso}
{\sc Siemon, D.}
\newblock Queueing in the linux network stack.
\newblock {\em Linux Journal 2013}, 231 (2013), 2.

\end{thebibliography}

\end{document}